 \documentclass[10pt,journal,compsoc]{IEEEtran}
\ifCLASSOPTIONcompsoc
  \usepackage[nocompress]{cite}
\else
  \usepackage{cite}
\fi
\ifCLASSINFOpdf
  \usepackage[pdftex]{graphicx}
\else
  \usepackage[dvips]{graphicx}
\fi
\usepackage{amsmath}
\ifCLASSOPTIONcompsoc
 \usepackage[caption=false,font=footnotesize,labelfont=sf,textfont=sf]{subfig}
\else
 \usepackage[caption=false,font=footnotesize]{subfig}
\fi
\usepackage{stmaryrd}

\usepackage{multirow}
\usepackage{amssymb}
\usepackage{booktabs}
\usepackage{xcolor}

\newcommand{\change}[1]{{\color{blue}#1}}
\renewcommand{\change}[1]{#1}
\newcommand{\rechange}[1]{{\color{blue}#1}}
\renewcommand{\rechange}[1]{#1}
\usepackage{hyperref}
\begin{document}
%
% paper title
% Titles are generally capitalized except for words such as a, an, and, as,
% at, but, by, for, in, nor, of, on, or, the, to and up, which are usually
% not capitalized unless they are the first or last word of the title.
% Linebreaks \\ can be used within to get better formatting as desired.
% Do not put math or special symbols in the title.

\title{Reducing Data Complexity using Autoencoders with Class-informed Loss Functions\thanks{This paper has been accepted for publication by IEEE TPAMI, please find the full article at \href{https://doi.org/10.1109/TPAMI.2021.3127698}{10.1109/TPAMI.2021.3127698}.\\\textcopyright 2021 IEEE. Personal use of this material is permitted. Permission from IEEE must be obtained for all other uses, in any current or future media, including reprinting/republishing this material for advertising or promotional purposes, creating new collective works, for resale or redistribution to servers or lists, or reuse of any copyrighted component of this work in other works.}}

%
%
% author names and IEEE memberships
% note positions of commas and nonbreaking spaces ( ~ ) LaTeX will not break
% a structure at a ~ so this keeps an author's name from being broken across
% two lines.
% use \thanks{} to gain access to the first footnote area
% a separate \thanks must be used for each paragraph as LaTeX2e's \thanks
% was not built to handle multiple paragraphs
%
%
%\IEEEcompsocitemizethanks is a special \thanks that produces the bulleted
% lists the Computer Society journals use for "first footnote" author
% affiliations. Use \IEEEcompsocthanksitem which works much like \item
% for each affiliation group. When not in compsoc mode,
% \IEEEcompsocitemizethanks becomes like \thanks and
% \IEEEcompsocthanksitem becomes a line break with idention. This
% facilitates dual compilation, although admittedly the differences in the
% desired content of \author between the different types of papers makes a
% one-size-fits-all approach a daunting prospect. For instance, compsoc 
% journal papers have the author affiliations above the "Manuscript
% received ..."  text while in non-compsoc journals this is reversed. Sigh.

\author{David~Charte,
  Francisco~Charte,~\IEEEmembership{Member,~IEEE,}
  and~Francisco~Herrera,~\IEEEmembership{Senior~Member,~IEEE}%
  \IEEEcompsocitemizethanks{%
    \IEEEcompsocthanksitem D. Charte is with the Computer Science and A.I. Dept., University of Granada, 18071 Granada, Spain. E-mail: fdavidcl@ugr.es.%\protect\\
    \IEEEcompsocthanksitem F. Charte is with the Computer Science Dept., University of Ja{\'e}n, 23071 Ja{\'e}n, Spain. E-mail: fcharte@ujaen.es.%\protect\\
    \IEEEcompsocthanksitem F. Herrera is with the Computer Science and A.I. Dept., University of Granada, 18071 Granada, Spain, and with the Faculty of Computing and Information Technology, King Abdulaziz University, 21589 Jeddah, Saudi Arabia. E-mail: herrera@decsai.ugr.es.%\protect\\
  }
}
\ifCLASSOPTIONpeerreview
  %\markboth{IEEE Transactions on Pattern Analysis and Machine Intelligence}%
  %{Reducing Data Complexity using Autoencoders with Class-informed Loss Functions}
\else
  %\markboth{IEEE Transactions on Pattern Analysis and Machine Intelligence}%
  %{Charte \MakeLowercase{\textit{et al.}}: Reducing Data Complexity using Autoencoders with Class-informed Loss Functions}
\fi
% The only time the second header will appear is for the odd numbered pages
% after the title page when using the twoside option.
% 
% *** Note that you probably will NOT want to include the author's ***
% *** name in the headers of peer review papers.                   ***
% You can use \ifCLASSOPTIONpeerreview for conditional compilation here if
% you desire.

% The publisher's ID mark at the bottom of the page is less important with
% Computer Society journal papers as those publications place the marks
% outside of the main text columns and, therefore, unlike regular IEEE
% journals, the available text space is not reduced by their presence.
% If you want to put a publisher's ID mark on the page you can do it like
% this:
%\IEEEpubid{0000--0000/00\$00.00~\copyright~2015 IEEE}
% or like this to get the Computer Society new two part style.
%\IEEEpubid{\makebox[\columnwidth]{\hfill 0000--0000/00/\$00.00~\copyright~2015 IEEE}%
%\hspace{\columnsep}\makebox[\columnwidth]{Published by the IEEE Computer Society\hfill}}
% Remember, if you use this you must call \IEEEpubidadjcol in the second
% column for its text to clear the IEEEpubid mark (Computer Society jorunal
% papers don't need this extra clearance.)

% use for special paper notices
%\IEEEspecialpapernotice{(Invited Paper)}

% for Computer Society papers, we must declare the abstract and index terms
% PRIOR to the title within the \IEEEtitleabstractindextext IEEEtran
% command as these need to go into the title area created by \maketitle.
% As a general rule, do not put math, special symbols or citations
% in the abstract or keywords.
\IEEEtitleabstractindextext{%
  \begin{abstract}
    Available data in machine learning applications is becoming increasingly complex, due to higher dimensionality and difficult classes. There exists a wide variety of approaches to measuring complexity of labeled data, according to class overlap, separability or boundary shapes, as well as \rechange{group morphology}. Many techniques can transform the data in order to find better features, but few focus on specifically reducing data complexity. Most \rechange{data transformation} methods mainly treat the dimensionality aspect, leaving aside the available information \rechange{within} class labels which can be useful when classes are somehow complex.

    This paper proposes \change{an} autoencoder-based approach to complexity reduction, using class labels in order to inform the loss function about the adequacy of the generated variables. \change{This leads to three different new feature learners}, Scorer, Skaler and Slicer. They are based on Fisher's discriminant ratio, the Kullback-Leibler divergence and least-squares support vector machines, respectively. \rechange{They can be applied as a preprocessing stage for a binary classification problem.} A thorough experimentation across a collection of 27 datasets and a range of complexity and classification metrics shows that class-informed autoencoders perform better than 4 other popular unsupervised feature extraction techniques, especially when the final objective is using the data for a classification task.
  \end{abstract}

  % Note that keywords are not normally used for peerreview papers.
  % \begin{IEEEkeywords}
  % Computer Society, IEEE, IEEEtran, journal, \LaTeX, paper, template.
  % Data complexity, feature learning, neural networks, 
  % \end{IEEEkeywords}
}

% make the title area
\maketitle

% To allow for easy dual compilation without having to reenter the
% abstract/keywords data, the \IEEEtitleabstractindextext text will
% not be used in maketitle, but will appear (i.e., to be "transported")
% here as \IEEEdisplaynontitleabstractindextext when the compsoc 
% or transmag modes are not selected <OR> if conference mode is selected 
% - because all conference papers position the abstract like regular
% papers do.
\IEEEdisplaynontitleabstractindextext
% \IEEEdisplaynontitleabstractindextext has no effect when using
% compsoc or transmag under a non-conference mode.

% For peer review papers, you can put extra information on the cover
% page as needed:
% \ifCLASSOPTIONpeerreview
% \begin{center} \bfseries EDICS Category: 3-BBND \end{center}
% \fi
%
% For peerreview papers, this IEEEtran command inserts a page break and
% creates the second title. It will be ignored for other modes.
\IEEEpeerreviewmaketitle

\IEEEraisesectionheading{\section{Introduction}\label{sec.intro}}
% Computer Society journal (but not conference!) papers do something unusual
% with the very first section heading (almost always called "Introduction").
% They place it ABOVE the main text! IEEEtran.cls does not automatically do
% this for you, but you can achieve this effect with the provided
% \IEEEraisesectionheading{} command. Note the need to keep any \label that
% is to refer to the section immediately after \section in the above as
% \IEEEraisesectionheading puts \section within a raised box.

% The very first letter is a 2 line initial drop letter followed
% by the rest of the first word in caps (small caps for compsoc).
% 
% form to use if the first word consists of a single letter:
% \IEEEPARstart{A}{demo} file is ....
% 
% form to use if you need the single drop letter followed by
% normal text (unknown if ever used by the IEEE):
% \IEEEPARstart{A}{}demo file is ....
% 
% Some journals put the first two words in caps:
% \IEEEPARstart{T}{his demo} file is ....

A classical obstacle in the field of data science is obtaining data of sufficient quality in order to extract the desired knowledge. The process of learning a model can be notably hindered by data presenting very common traits such as noise \cite{xiong2006enhancing}, outliers \cite{aggarwal2015outlier}, high dimensionality \cite{AYESHA202044} or complex class boundaries \cite{ho2000measuring}. This results in long periods of time spent cleaning and preprocessing data \cite{garcia2015data} before the actual data mining step can even begin.
Although data cleaning techniques can be of good use in order to identify and filter out noise, outliers and missing data, other aspects can be trickier to solve.

Many real world situations can be modeled as supervised classification problems, those where each instance belongs to one of several classes, and the objective is to learn from the observed data from each class in order to automatically assign the corresponding class labels to new, unobserved instances. Some examples of classification problems are text categorization~\cite{Yang1999ARO}, spam filtering~\cite{Dada2019MachineLF}, object recognition in images~\cite{Russakovsky2015ImageNetLS} and automatic interpretation of medical data to facilitate diagnostics~\cite{deo2015machine}. \rechange{Many of these problems correspond to the simplest case, binary classification, where there are only two categories.}

\begin{figure*}[hbtp]\centering

  \hfill\includegraphics[width=.3\textwidth,trim={1cm 2cm 0 0},clip]{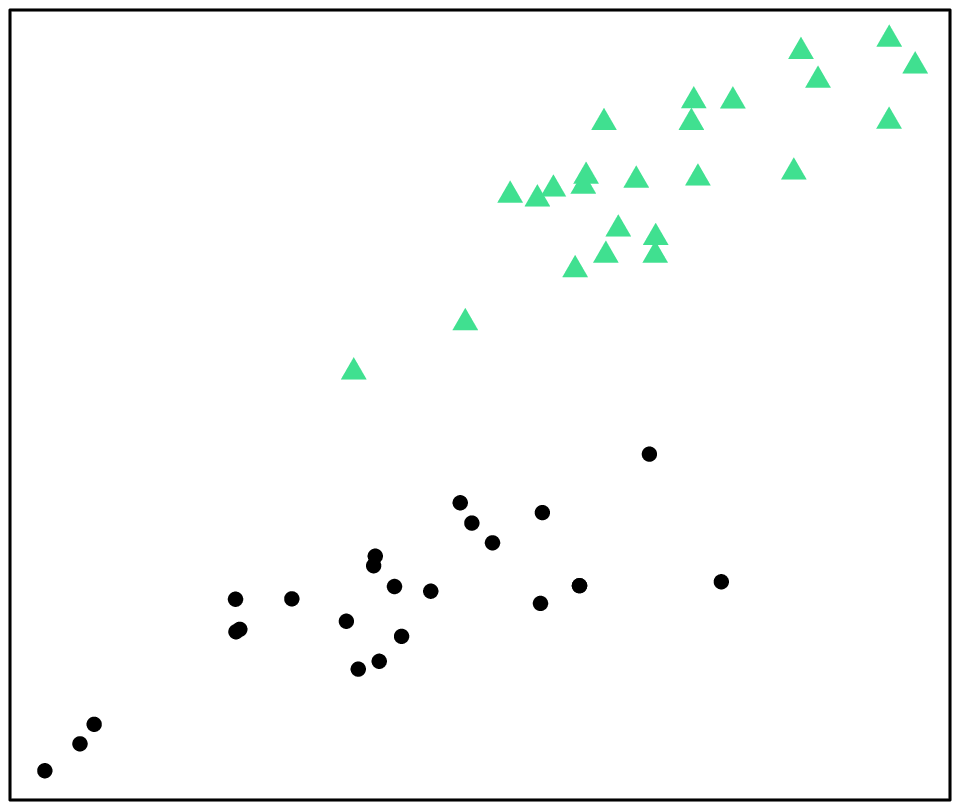}%
  \hfill\includegraphics[width=.3\textwidth,trim={1cm 2cm 0 0},clip]{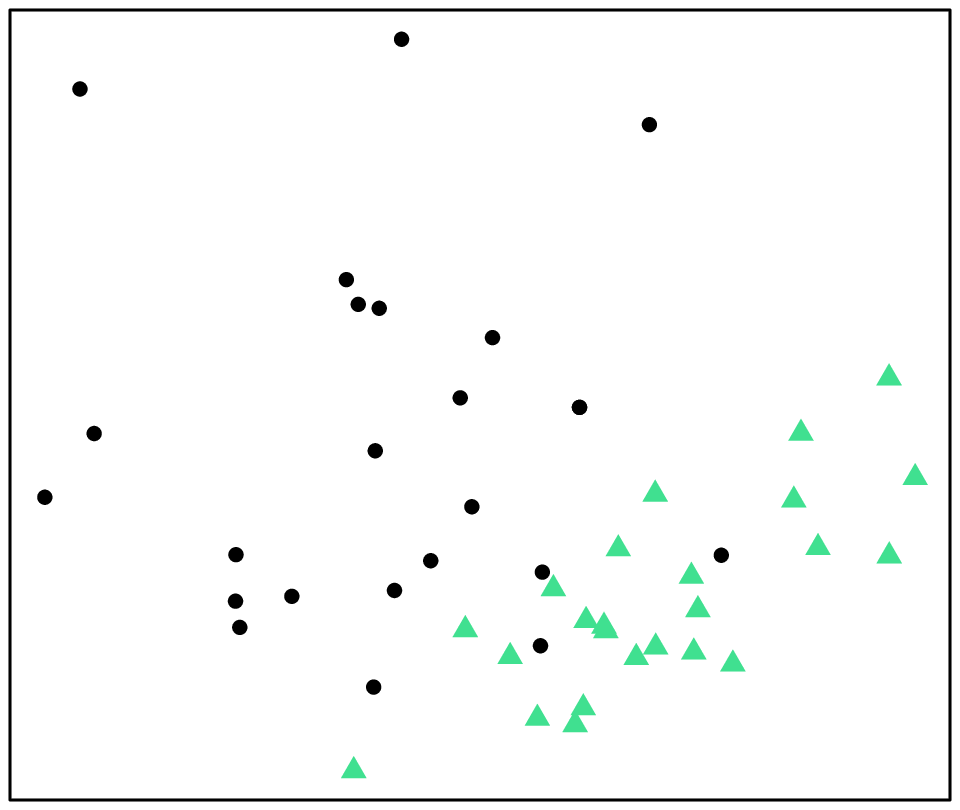}%
  \hfill\includegraphics[width=.3\textwidth,trim={1cm 2cm 0 0},clip]{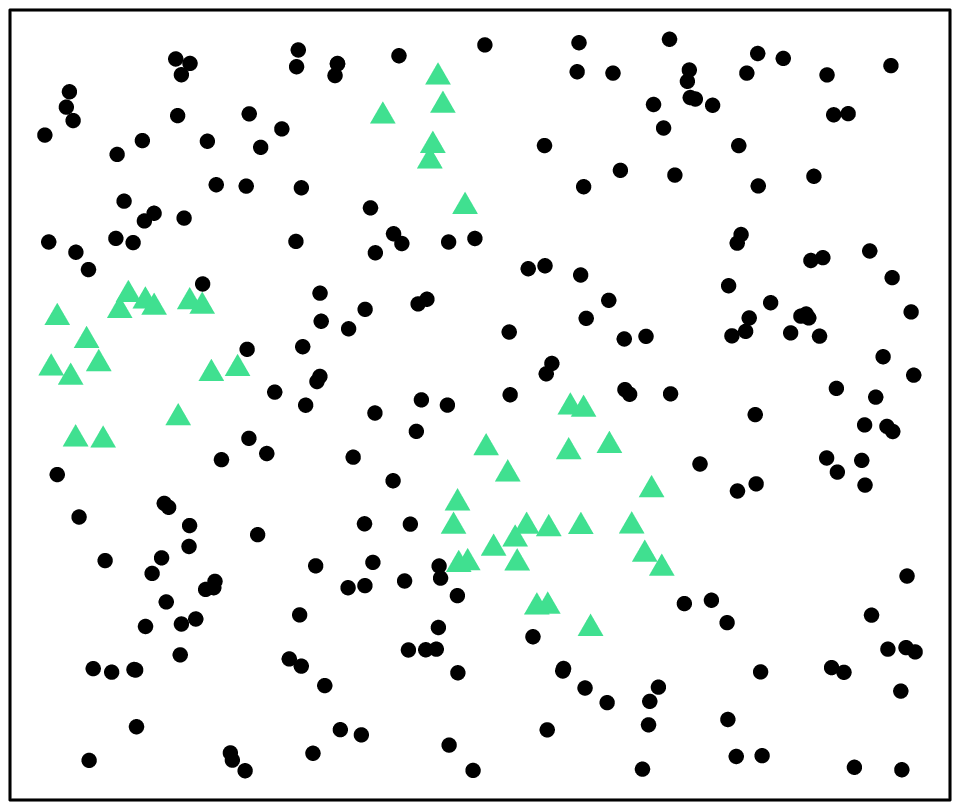}\hfill%
  \caption{\label{fig.boundaries}Different situations relating to class complexity. The graph on the left shows separable classes, the middle one is an example where classes are not separable and the one on the right shows separable classes with complex boundaries (small disjuncts).}
\end{figure*}

One of the type of issues that is very commonly overlooked in classification problems is the complexity of data~\cite{ho2000measuring,lorena2019complex}. Consider a clean dataset with no presence of errors or abnormalities. There can still be aspects related to the geometrical shapes and overlap among classes which can hinder the performance of a learning technique. For example, there could be no separability between classes, or even regions of the feature space with a mix of instances from different classes. In cases where separability is achievable, boundaries can present complex shapes that can be difficult for a parameterized model to fit. Figure~\ref{fig.boundaries} illustrates some of these cases, more concretely, one where the features do not allow to separate the classes, and another where class boundaries are difficult to model due to there being several small groups from one class, sometimes known as small disjuncts~\cite{weiss1995learning},  within a group of the other.

An additional hindrance that frequently occurs in data mining scenarios is associated to the representation of instances, to the features themselves \cite{DimRecComparative}. These can be typically seen as observed outcomes of underlying factors that cause them and, as a result, are not always the ideal representation of the data. This can depend on the objective task and the learning method to be used. For example, a pixel-based representation for images can be ideal for a convolutional neural network to perform classification, but may be difficult for a lazy learning method to process.

When data have some kind of complexity, it can affect the performance of machine learning methods and these are usually not able to overcome the issue by themselves. Instead, \rechange{a preprocessing step can transform data aiming to find} a better representation which makes it easier to categorize points. Operating with features for this purpose is a task known as feature extraction or feature learning. There exists a wide variety of approaches to feature extraction \cite{bengio2013representation}, including linear transformations, manifold learning and neural network-based models. However, very few of them take class complexity into account and, as a result, extracting quality features with a specific strategy to reduce this complexity is still an important challenge.

In particular, autoencoders (AEs) \cite{charte-tutorial} are neural networks specifically designed to extract features from the data. These are typically trained to reconstruct the input at its output, feeding the data through several layers which impose some kind of restriction or bottleneck in order to find more appropriate representations along the way. AEs can also be easily restricted or adapted in order to promote certain kinds of transformations and encodings, for example, finding sparse variables which only take high values for a small number of instances \cite{ng2011sparse}.

\change{This work makes use of the well-known technique for regularizing the behavior of an AE, applied in this case to achieve class complexity reduction. Aiming to transform features onto a more useful space with special attention to class complexity, three concrete models that use different criteria are proposed. The bases for these are}: Fisher's discriminant ratio, the Kullback-Leibler divergence (KLD) and least-squares support vector machines (LSSVMs). The new models have been tested against well-established feature extraction methods \rechange{within a binary classification pipeline}.

In summary, the main contributions of this paper are the following:
\begin{itemize}
  \item New AE-based models able to learn from input features as well as binary class labels, specifically the following three variants:
        \begin{itemize}
          \item Scorer, a model which enables separability among classes by means of the Fisher's discriminant ratio.
          \item Skaler, a model that receives feedback from the KLD and can thus provide features where positive and negative instances belong to very different distributions.
          \item Slicer, an extended AE using a LSSVM in order to simultaneously evaluate a simple linear classifier and assess the adequacy of the new features for classification.
        \end{itemize}
  \item A thorough experimentation across 27 cases and 11 evaluation measures, focusing on different complexity rates and classification performance, and against 4 other well-known feature extraction methods.
  \item A comparison between the most interpretable complexity metrics and several evaluation metrics for classifiers, revealing which of the complexity metrics are better predictors of classification performance.
\end{itemize}

As an important conclusion after the experimental analysis of the newly proposed models, we must point out that they can be trained to generate better features for the purposes of classification than other popular feature extraction methods.

The rest of this paper is organized as follows. Section~\ref{sec.sota} describes the current state with respect to available complexity measures and techniques to overcome complexity in data. Next, Section~\ref{sec.aecompred} introduces our proposals and provides all the details about their inner workings. Section~\ref{sec.setup} explains the details of the experimentation process, while Section~\ref{sec.experiments} discusses the results. Lastly, conclusions and final comments are provided in Section~\ref{sec.comments}.

% \begin{itemize}
%   \item Why is data quality important
%   \item What is data complexity (what is classification?)
%   \item What complex data looks like, what simple data look like (how is it measured?)
%   \item Why mining models do not solve complexity, how are they hindered by complexity
%   \item What are the current approaches to solve data complexity
%   \item What is feature learning, what are autoencoders
%   \item What is the proposal of this work
% \end{itemize}

\section{State of the art in complexity reduction}\label{sec.sota}

A dataset can present many different problems that may drive it to be considered difficult or complex to classify. Initially, a possible measure of complexity could be the error rate of the classifier itself. However, the objective of this work is identifying difficult datasets to treat them before learning a classifier. For this reason, we rely on other metrics which aim to characterize the complexity of supervised problems.

\subsection{Sources of difficulty}

Ho and Basu \cite{basubook} identify three possible sources of difficulty: (1) class ambiguity, (2) boundary complexity and (3) sample sparsity and feature space dimensionality. The first applies to the circumstances where classes cannot be distinguished using the given features, either because they provide insufficient knowledge about the problem or because the classes are not well defined. The second source refers to the situation where classes are interleaved or not easily separable. In these cases, the complexity can be measured attending to class overlap and class separability as well as geometry, topology and density of manifolds. The last category covers issues with the structure of the sample, whether it is complete enough and the amount of variables the classifier needs to work with.

% In this work we focus on treating some of these sources of difficulty, namely: class overlap, class separability and feature space dimensionality.

\subsection{Complexity measures}\label{sec.compmeas}

In order to quantitatively assess how complex a dataset is, a wide variety of complexity metrics have been proposed over the years \cite{ho2000measuring,lorena2019complex}. The following sections briefly describe the most relevant approaches to measure complexity, paying special attention to the metrics that will be \rechange{applied} throughout the experimentation. Each metric is abbreviated according to the original nomenclature in \cite{ho2000measuring} and \cite{lorena2019complex}.

\subsubsection{Class overlap}

Geometrical complexity in a dataset can be characterized in several ways. One approach is to measure the overlap in feature values among different classes. Each feature can be assessed as to how much it contributes to distinguishing the classes. In this case, measures usually focus on binary problems. \change{The following measures follow this approach}:

\paragraph*{Maximum Fisher's discriminant ratio (F1)}
Fisher's discriminant ratio is a measure of class overlap, based on the simplest statistics for a distribution, mean and standard deviation. \iffalse It is defined for each feature as in Equation~\ref{eq.fdr}.
  \begin{equation}\label{eq.fdr}\operatorname{dr}(f_i) = \frac{\left(\mu_i^{+}-\mu_i^{-}\right)^2}{\left(\sigma_i^{+}\right)^2+\left(\sigma_i^{-}\right)^2}~,\end{equation}
  where $\mu_i^{+}$ and $\mu_i^{-}$ are the average values for the $i$-th feature in each class, and $\sigma_i^{+}$ and $\sigma_i^{-}$ are the corresponding standard deviations. This ratio is computed for each variable and the maximum is chosen to characterize the dataset: $\operatorname{F1}=max_{i=1}^m \operatorname{dr}(f_i)$. \fi  \change{Higher values of this metric mean lower levels of overlap. The maximum over all features is taken as a measure of the class separation in a dataset.}
%  \paragraph*{Volume of overlapping region (F2)}
%  This measure considers the overlap between the bounding boxes of both classes, represented by a product of the overlapped proportion of the range of each feature.
%  \iffalse
% \begin{align}
%   \operatorname{F2} &= \prod_i^m \frac{\operatorname{overlap}(f_i)}{\operatorname{range(f_i)}} \\
%   &= \prod_i^m \frac{\max(0, \min\max(f_i)-\max\min(f_i))}{\operatorname{\max\max(f_i)-\min\min(f_i)}}~,\mbox{ where}
% \end{align}
%   \begin{align}
%     \min\max(f_i) &= \min\left(\max\left(f_i^{+}\right), \max\left(f_i^{-}\right)\right)\\
%     \max\min(f_i) &= \max\left(\min\left(f_i^{+}\right), \min\left(f_i^{-}\right)\right)\\
%     \min\min(f_i) &= \min\left(\min\left(f_i^{+}\right), \min\left(f_i^{-}\right)\right)\\
%     \max\max(f_i) &= \max\left(\max\left(f_i^{+}\right), \max\left(f_i^{-}\right)\right),
% \end{align}
% $f_i^+$ is the sequence of values of the $i$-th feature for the positive class and $f_i^-$ is analogous for the negative class.\fi
% The resulting metric is higher when the overlap among classes is higher. The main problem with this metric is that it assumes that there is no more than one cluster of each class: when the opposite happens, a cluster of positive instances could be in the middle of two clusters of the negative class without overlap but the metric would find maximum overlap volume.
\paragraph*{Maximum feature efficiency (F3)} Feature efficiency is calculated as the proportion of examples that can be unambiguously classified by a simple threshold, that is, they lie outside an overlapping region. This rate gives an idea of the usefulness of a given feature when attempting to classify every instance in the dataset. The maximum of this ratio across all features is known as F3.
\iffalse\begin{equation}
  \operatorname{F3}=\max_{i=1}^m\frac{n-n_o(f_i)}{n}~\mbox{, where}
\end{equation}\begin{equation}
  n_o(f_i)= \sum_{j=1}^n\left\llbracket \max\min(f_i) < x_{j_i} < \min\max(f_i)\right\rrbracket
\end{equation}
and $\left\llbracket\dots\right\rrbracket$ is the Iverson bracket, which evaluates to 1 if the inner statement is true, and to 0 otherwise. As a result, $n_o$ counts the number of values observed for the $i$-th feature that fall within its overlapping region. \fi

\subsubsection{Class separability and nonlinearity}

Instead of measuring the importance of the overlapping regions in features, an alternative approach is to look for complexity of the boundary separating classes, that is, its ability to actually isolate both classes and its nonlinearity. Several measures have been developed regarding the shape and separation degree of classes.

%\paragraph*{Linear classifier objective function (L1)} The linear classifier proposed by Smith \cite{smith1968pattern} can help  measure class linear separability. In particular, the value of the objective function that has to be minimized can be also used as a measure of the level of separability or nonlinearity: the higher L1 is, the more difficult it is for a linear classifier to separate instances from both classes.
\paragraph*{Linear classifier error (L2)} Linear separability of classes is the core of a branch of classification methods, support vector machines (SVM) \cite{steinwart2008support}. In its simplest form, a SVM is a binary classifier that attempts to find the hyperplane which best separates both classes. Its training error can be used as a metric to characterize the separability (or lack thereof) of a dataset.
\paragraph*{Linear classifier nonlinearity (L3)} Describing the shape of the regions occupied by each class can also contribute to learning about the complexity of the data.  In particular, this measure tackles nonlinearity, i.e. the smoothness of the decision boundary of a classifier, which can be detected by interpolating pairs of points of the same class to extract a test set and computing the classification error for this new set.

\subsubsection{Neighborhoods and morphology}

The previous traditional measures for data complexity come from a statistical or geometrical point of view. Other metrics look at how instances are located around each other, so they study local behavior instead of global properties.

\paragraph*{1-NN classifier error (N3)} Similarly to the L1-L3 measures, which make use of a simple classifier in order to measure complexity, this metric performs a leave-one-out validation of a nearest neighbor classifier, that is, it checks the class of every instance according to the nearest one, and measures complexity as the error rate obtained.
%\paragraph*{Space covering by $\varepsilon$-neighborhoods (T1)} A system can be defined to group points from the same class in neighborhoods that are built progressively by advancing some distance $\varepsilon$ as many times as possible without including points of the opposite class.

Recently, some new metrics have been proposed that attempt to describe data complexity from the perspective of data morphology \cite{pascual2020revisiting}. These are based on the Pure Class Cover Catch Digraph (P-CCCD) classification method \cite{pcccd}.

P-CCCD creates a collection of balls that cover the feature space so that each ball only contains points from the same class. The process consists in choosing a ball so that it is centered in a point of the target class and is the largest possible ball that does not any point of the other class. This is repeated until all points of that class are in at least one ball, producing a cover which is not necessarily optimal but is a good approximation.

Morphology-based complexity metrics are inspired by this algorithm in the sense that they look for a ball cover of all points where balls only contain points from one class, and then perform some computations according to the number of balls created. The main metrics are as follows:

\paragraph*{Total number of balls ($\textit{ONB}_{\text{tot}}$)} This measure counts the total number of balls required to produce the cover. If $b^+$ balls are needed to cover all positive instances and $b^-$ are necessary \change{for} the negative points, it is calculated as \begin{equation}\textit{ONB}_{\text{tot}}=\frac{b^++b^-}{n}~.\end{equation}
\paragraph*{Average number of balls ($\textit{ONB}_{\text{avg}}$)}  It averages the amount of balls used to cover the points of each class. In a binary classification environment, the definition would just be the sum of the balls-to-points ratios, divided by 2: \begin{equation}\textit{ONB}_{\text{tot}}=\frac{\frac{b^+}{n^+}+\frac{b^-}{n^-}}{2}~.\end{equation}

These metrics can turn into a very general way of describing the geometrical complexity of the classes, since the shape of the balls depends on the distance chosen (e.g. Euclidean, Manhattan or the maximum distance). The mechanism for covering the feature space attempts to use as few balls as possible to cover all points. If the dataset can be covered by a few large balls, then its complexity will be low, but if many small balls are needed, it means that many little clusters of different classes are near each other, and the complexity is thus high. Both $\textit{ONB}_{\text{tot}}$ and $\textit{ONB}_{\text{avg}}$ are, as a result, higher the more complex the data is. The difference between them is that $\textit{ONB}_{\text{avg}}$ gives the same weight to all classes, while $\textit{ONB}_{\text{tot}}$ does not distinguish classes but gives the same weight to all instances.

\subsubsection{Feature space dimensionality}

One of the main issues that occur in many datasets and has been tackled from many perspectives is dimensionality. Dimensionality refers to the number of variables where each instance takes values. High dimensionality has long been considered a problem for classification algorithms, known as curse of dimensionality \cite{aggarwal2001surprising}. It is not directly related to the way classes interact with each other, but a high number of features can hinder the performance of a classifier with a dataset that is otherwise not considered complex, due to the fact that most distance metrics lose meaning when measuring across many variables.

Dimension can be measured in absolute terms, but the complexity that derives from it is also related to the number of instances in the dataset. Two problems with the same number of features are not equally complex if the first one has 10 times more instances than the other. As a result, \change{an instances-to-features rate (T2) can be considered a complexity metric that can give a better account of this relation}.

%\paragraph*{Dimension ($d$)} This is just the number of input attributes in a dataset.
%\paragraph*{Instance-feature rate (T2)} If there are $d$ variables describing the instances in the dataset, this rate is computed as \begin{equation}\operatorname{T2}=\frac{n}{d}~.\end{equation}

\subsection{Other models for complexity}

When trying to reduce the complexity present in a dataset, one can take complexity measures into account for evaluation purposes, and use other ways of modeling complexity when training and performing data transformations. For instance, considering that each class presents different distributions across each variable, some similarity or dissimilarity metrics for distributions could be used.

The Kullback-Leibler divergence is a well-known measure of how a distribution differs from another one, it is asymmetric as it usually compares a distribution coming from data with a distribution representing a model or theory. If these are defined on a discrete probability space $\mathcal X$, then the divergence is formulated as
\begin{equation}D_{\operatorname{KL}}(p\Vert q)=\sum_{x\in\mathcal X} p(x)\log\frac{p(x)}{q(x)}~.\end{equation}

This quantity could provide an intuition on how two distributions are overlapping or separated. It is higher the more different the distributions are. One way to retrieve a symmetric value out of it is to add the Kullback-Leibler divergence of the distributions in reverse order: \(D_{\operatorname{KL}}(p\Vert q)+D_{\operatorname{KL}}(q\Vert p)\).
For both measures, the chosen distribution when applying them to class separability could be a Bernoulli distribution for each feature in the encoding, so that their values are considered either high or low. If we model all features at the same time, a categorical distribution could be employed. Assuming a binary classification problem, we could measure the dissimilarity of the distribution corresponding to positive instances against the distribution of negative instances, which would provide a sense on how easy it is to differentiate them.

\subsection{Reducing complexity in datasets}\label{sec.methods}

There are several approaches to complexity reduction in datasets. This section provides a general overview of the different aspects that can be treated and techniques for doing so. %Most of them deal with the dimensionality of the data, but from different perspectives.

%%%%% Antes de la siguiente sección, no sé si como una subsección aquí o como sección independiente, habría que hacer una revisión del state-of-the-art de las propuestas existentes para reducción de complejidad, citando los papers más importantes con este tipo de propuestas

%\subsubsection{Dimensionality reduction}

Dimensionality of data has been one of the most diversely tackled issues. A multitude of methods exist in the literature, ranging from simple feature selection to nonlinear feature learning. A thorough review of all these can be found in \cite{garcia2015data}, but we enumerate and describe the main ones below.

However, there are other emerging methods that may be able to modify other aspects of the data and reduce complexity along the way. Some of those are distance metric learning methods.

\subsubsection{Feature selection}

Assuming that not every variable has the same relevance for the purposes of classification, an initial approach to dimensionality reduction can be to simply discard some of them, retaining only the ones that help the classifier the most. This process is known as feature selection \cite{garcia2015data}. Of course, there exist a plethora of criteria that can apply for this purpose.

\paragraph*{Filters} This variety of techniques is mostly founded on statistical and information theory measures, such as the joint mutual information, the conditional mutual information, the Kullback-Leibler divergence or minimum-reduncancy-maximum-relevance. The objective is to quantify the utility of each variable and keep only the most useful ones. In fact, some approaches take class separability into account as well \cite{zhang2013divergence,wang2008feature}.

\paragraph*{Wrappers} Another way of looking at feature selection is shaping it as an optimization problem, finding an adequate fitness function, typically the performance of a classifier, and making use of one of the many existing metaheuristics available, for instance, genetic algorithms, simulated annealing or particle swarm optimization, to name a few.

\paragraph*{Embedded methods} Some classifiers have built-in feature selection, so that they only look at the information provided by the most relevant variables. These are usually decision trees like C4.5 \cite{quinlan}.

\subsubsection{Linear feature extraction}

Another way of reducing the number of variables is to attempt to summarize most of the information of the original variables in a smaller set of new variables, which emerge as linear transformations of the original ones.

\paragraph*{Principal component analysis (PCA) \cite{PCA,PCABook}} PCA  is a well-studied technique that solves the problem of obtaining features which retain the maximum possible variance while being uncorrelated to each other. It also allows to recover the original data from the projected points while losing the minimum amount of information as measured by the mean squared error. %Once the adequate transformation, based on the eigenvalues of the data matrix, is found, one can project the original points to their corresponding ones by simple matrix multiplication. PCA in itself does not provide a lower amount of features, but one can usually retain most of the variance of the data by keeping only the first principal components and omitting the rest, achieving thus a more compact representation.

\paragraph*{Linear discriminant analysis (LDA) \cite{LDA}} This is a supervised method able to extract linear combinations of features which achieve good class separation. Under assumptions of normality, independence and homoscedascity, it can project the data onto a space consisting of new coordinates that best discriminate the classes. \change{Its} main drawback is that the number of resulting variables is completely determined by the number of classes. More specifically, for a problem with $c$ classes, LDA will output a space of $c-1$ linear combinations of the original variables. There is a recent generalization of LDA which claims to solve its stability issues and achieve better class separation through a maximum margin criterion \cite{li2006efficient}.

\paragraph*{Factor analysis \cite{PCAandFA}} This technique assumes, unlike PCA, that a series of hidden factors are generating the observed data by means of linear combinations. The number of underlying factors is lower than the number of observed variables, and they are assumed to have zero mean and unit covariance (i.e. the identity matrix). %The original data is recovered from the factors by linear combinations, where the coefficients of each term are known as loadings, and some level of stochastic error.

% DONE: completar estos dos

\subsubsection{Nonlinear feature extraction}

The most advanced methods for dimensionality reduction base their new variables on nonlinear transformations of the original ones.

A lot of these techniques can be grouped in a concept known as manifold learning, since they attempt to find structure for a manifold where most of the data lie, and thus transform each data point onto its coordinates on that manifold.

\paragraph*{Multidimensional scaling (MDS) \cite{borg2005modern}}
This is a classical methodology that has served as basis for several other algorithms as well. Its objective is to compute new coordinates for data points while preserving distances among them as faithfully as possible. Instead of having points as inputs, it only receives the pairwise distances themselves, and minimizes a loss function which helps the model obtain coordinates for each point, creating a space where the given distances are maintained. %There are several loss functions according to the type of MDS applied. For example, in metric MDS one would measure Euclidean distances among points and then minimize the stress function, which is a sum of squared errors.

\paragraph*{Isomap \cite{Isomap}} This method extends metric MDS in order to find coordinates that describe the actual degrees of freedom of the data while preserving distances among neighbors and geodesic distances between the rest of points. %The main difference between MDS and Isomap is that, while the Euclidean distance is usually employed for any pair of points in MDS, 
Isomap constructs a neighborhood graph where each edge is weighted according to the Euclidean distance among vertices\change{, then uses this} to compute geodesic distances instead of using straight lines. These new distances are potentially higher than the Euclidean but help capture more information about the manifold.

\paragraph*{Locally linear embedding (LLE) \cite{LLE}} The objective of LLE is similar to that of the previous techniques, but with a different approach to preserving the local structure. It finds a linear combination which describes \change{each point} from its neighbors. Once \change{its} coefficients have been computed, LLE optimizes the coordinates for a lower-dimensional space so that they fit the same expressions.

\paragraph*{t-stochastic neighbor embedding (t-SNE) \cite{van2008visualizing}} This is a technique specially oriented for visualization, so it finds specially attractive low-dimensional projections of the data. %The method for finding the embedding works somewhat differently than the rest, although there exists some similarities with LLE. 
It consists on assigning, for each pair of points, the probability that one point would choose the other as its nearest neighbor if neighbors are computed according to Gaussian distributions centered on each point. t-SNE then defines a low-dimensional mapping that tries to preserve these probability scores.

\paragraph*{Autoencoder networks (AE) \cite{charte-tutorial}} AEs are neural network models which reconstruct the input at their output, using some kind of bottleneck in between so as to learn useful information from the data. We explain AEs in further detail in Section~\ref{sec.aefund}.

% DONE terminar de explicar estos

\subsubsection{Distance metric learning}

Distance metric learning \cite{suarez2018tutorial} is an area of machine learning dedicated to learning distances from datasets. These distances are built to better represent the similarities and differences among examples than standard distances, such as the Euclidean distance.

In a supervised learning context, the problem of learning a distance can be formulated as follows:
\begin{align}
    & \operatornamewithlimits{arg min}_{d\in \mathcal D} l(d, S, D)\mbox{, where} \\
  S & =\{(x_i, x_j)\in \mathcal X\times \mathcal X\, :\, y_i=y_j\}                \\
  D & =\{(x_i, x_j)\in \mathcal X\times \mathcal X\, :\, y_i\neq y_j\}
\end{align}
and $l$ is a loss function that determines the fitness of a distance to describe the similarities and differences provided as sets $S$ and $D$.

Some of the dimensionality reduction methods mentioned above can be also seen as distance metric learning techniques, but there are more algorithms which can learn distances. Some of them are addressed at improving the performance of k-nearest neighbors, and others are based on information theory. Among the most relevant are: NCA \cite{goldberger2004neighbourhood}, LMNN \cite{weinberger2009distance}, NCMML \cite{NCMC} and NCMC \cite{NCMC}.

% DONE: completar o reducir

\subsubsection{Current limitations}
Many of the complexity reduction methods explained above tackle complexity only partially or from a limited perspective. In most of the cases, tha main focus is reducing dimensionality. This can help model data when good quality coordinates are found, but may discard useful information present in the class labels.

Some of the available methods consider class separability when selecting features \cite{zhang2013divergence,wang2008feature} but they do not generate new features, and can capture only a partial view of the whole feature space as a result.

In summary, there is an unexplored possibility of advanced feature extraction techniques which perform nonlinear transformations of variables in order to find spaces where classes are further away and easier to identify.

\section{Autoencoders for complexity reduction}
\label{sec.aecompred}

The objective of this work is to develop strategies that address data complexity in a more complete way, that is, working with transformations of all available features and incorporating class complexity measures to acquire information from the class labels. The result is a collection of models that are based on AEs because they are very versatile deep learning architectures, able to transform the data in diverse ways according to their loss function. Our hypothesis is that when the loss function takes data complexity into account, then the AE will have more information to work with and will generate better features than other feature extraction methods.

In order to provide an indication on data complexity to a loss function, it is necessary to look for computationally simple ways of calculating a penalty that points the training method in the right direction. This problem can be approached from several possible perspectives, including the integration of a complexity metric, a measure of distribution dissimilarity or a linear separation method.

This section details the theoretical underpinings of our proposals: Scorer, Skaler and Slicer. First, some basic notions of AEs help establish a starting point for these new models. Next, the added penalties for Scorer and Skaler are explained. Lastly, the necessary modifications and computations needed for Slicer are shown as well.

% motivar los modelos, dar visión general

\subsection{Autoencoder fundamentals}
\label{sec.aefund}

A neural AE \cite{charte-tutorial,bengio2012deep} is generally a symmetrical neural network trained to reconstruct the inputs at its output. The composition of layers up to the middle one computes a new representation of the input data where some traits may be induced: lower dimension, sparsity, or robustness against noise, for example. The feature transformation is learned by means of a training process that optimizes the reconstruction error as well as, potentially, other penalties allowing the introduction of those specific aspects.

\begin{figure}[ht]
  \centering
  \includegraphics[width=.25\textwidth]{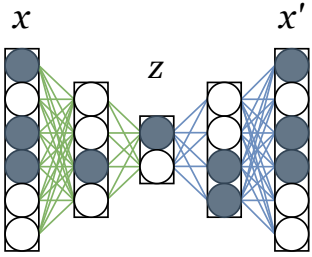}
  \caption{\label{fig.basicae}The essential structure of an AE implemented as a fully connected feed-forward neural network, composed of an encoder $f$ and a decoder $g$. The training loss of this model is measured as the distance $d$ between the input $x$ and its reconstruction $x'=(g\circ f)(x)$.}
\end{figure}

A simple AE models the reconstruction problem as a deterministic function given by the composition of an encoder $f$ and a decoder $g$. When an instance is feeded to the model, the encoder transforms it to a vector located within the encoding space, and the decoder maps this vector to the original feature space.

Consider the diagram in Fig.~\ref{fig.basicae}. During training, mini-batches of samples are propagated through the network. The AE is evaluated according to the average distance between original and reconstructed samples. Its weights are iteratively modified in order to minimize this distance. There are two typical dissimilarity metrics for this purpose:

\begin{itemize}
  \item Mean squared error: defined as the average of squared errors. \change{If $x$ and $x'$ are a training sample and its reconstruction, it is expressed as:} \begin{equation}\mathcal L(x,x')=\frac 1 n \sum_{i=1}^n(x_i - x_i')^2\end{equation}
  \item Cross entropy: this measure is effective when modeling data where values lie in the $[0,1]$ interval, since it is usually implemented as the cross entropy of two Bernoulli distributions. The formulation is as follows: \begin{equation}\mathcal L(x,x')= -\sum_{i=1}^n x_i\log x_i'  + (1-x_i)\log(1-x_i') \end{equation}
\end{itemize}

In general, any kind of measure that indicates the difference among two data points of the same type can be used. For certain types of structured data, such as images or sequences, specific reconstruction errors may also apply. For instance, a perceptual loss \cite{johnson2016perceptual} can be very fitting for image reconstruction, since it \change{focuses more in the  appearance of the image instead of} trying to accurately recover each individual pixel, which can lead to softer and blurrier images.

Once one of these dissimilarity metrics is chosen, the loss function of the AE can be defined:
\begin{equation}
  J(\theta;S)=\sum_{(x,y)\in S} \mathcal L(x, (g\circ f)(x))~,
\end{equation}
where $\theta$ holds the parameters of the network, and thus determines $f$ and $g$, and $S$ is a set of training instances.

Diverse kinds of regularizations can be applied to the loss function with the objective of adjusting the behavior of the AE\change{, such as sparsity, contraction or variational inference}. Each of these result in a slightly different AE variant with its own applications. \iffalse The most popular AE variants are the following

  \begin{itemize}
    \item Sparse AE \cite{ng2011sparse}: Introduces a small penalty term which forces the model to prevent many high activations in the encoding for each input data point, resulting in sparse codifications, where most units take the value 0.
    \item Contractive AE \cite{rifai_contractive_2011}: This model, which minimizes the Frobenius norm of the Jacobian of the encoder, facilitates encodings where neighborhood structure is preserved from the inputs, although global structure is transformed. The objective is not dissimilar to that of manifold learning methods such as LLE and Isomap.
    \item Variational AE \cite{VariationalAE}: The encoder in this model is not a deterministic function but a stochastic one, it maps each input instance onto a probability distribution, which is then sampled to obtain the decodification. The resulting encoding space can be sampled smoothly, generating new unobserved examples from the original data distribution.
    \item Denoising AE \cite{vincent_stacked_2010}: This approach is unlike the previous, since it does not modify the loss function but the input data, introducing artificial noise that the AE attempts to recover from and eliminate. This process help build more robust features that resist small perturbations in the inputs.
  \end{itemize}

\fi Although these and several other regularizations help build better feature spaces, to the best of our knowledge there is no AE variant focusing on enabling class separability or reducing data complexity yet.

AEs are generally trained with common neural network optimizers, such as stochastic gradient descent~\cite{robbins1951stochastic} or Adam~\cite{adam}. \change{They decide how to update the parameters in an iterative process which computes the gradient of the loss function via backpropagation~\cite{backprop}.}

\subsection{\change{Regularizing autoencoders with label information}}

As described above, a basic autoencoder is trained using a loss function which evaluates the distance between the input feature vector and its reconstruction through the network. It is clear, by its definition, that instance labels are not used at all to compute the loss function, nor does the AE receive this information as input. This has its advantages and shortcomings. A benefit is that one may train AEs using unlabeled data and obtain valuable knowledge as a result. This allows for their use in several widespread applications \cite{CHARTE202093}, such as anomaly detection, data denoising, synthetic instance generation and semantic hashing.

One possible drawback, when applying AEs to classification problems, is that they will extract features that may or may not help distinguish the classes, since they are not provided with the labels. However, this is not applicable to all AE-based models, since some of them can take the class label into account when computing the encoded representation, either directly as an input layer to the network, or indirectly, by informing the loss function.

A common way to modify the behavior of the training process and improve the solutions is to add a penalty term $\Omega$ to the loss function promoting certain aspects of the encoding or reconstruction mappings. This penalty may be dependent on the weights of the network or the resulting codes. It is added with a weight coefficient $\lambda$ in order to adjust its importance with respect to the standard reconstruction error:

\begin{equation}J\left(\theta;S\right)=\sum_{(x,y)\in S}\mathcal L\left(x,\left(g\circ f\right)\left(x\right)\right) + \lambda\Omega\left(\theta;S\right)\end{equation}

For instance, one well-known regularization consists in penalizing high levels of simultaneous activations within the codes. This, usually called a sparsity regularization \cite{ng2011sparse}, helps maintain a low number of active neurons in the encoding for each sample.

In our case, the objective is that the resulting feature transformation helps better separate different classes, so the loss function should receive some kind of label information in order to be able to learn from it. The procedure can thus be similar to a penalty modification, but using the class label within the penalty term $\Omega$.

There could be numerous ways of analyzing the relation of codes and classes. For example, trying to optimize a complexity measure or maximizing the difference among class distributions, as well as wrapping a simple classifier so as to assess the quality of the features. These are the main ideas behind our three proposals:
\begin{itemize}
  \item Scorer, an AE model with a Fisher's discriminant ratio-based penalty. Its objectives may be collaborating or in opposition, but it needs to find a balance between good instance reconstructions and low class overlap.
  \item Skaler, an AE using the KLD to separate class distributions. The encodings are modeled as a categorical distribution and the model attempts to maximize the divergence among the distribution of positive instances and that of negative instances. This should draw them apart from each other.
  \item Slicer, an AE which internally trains a linear least-squares support vector machine. The internal classifier need not be perfect, but it helps the model analyze how easy it is to classify the instances using the generated features. The objective, in this case, is to maximize the linear separation of both classes.
\end{itemize}
Along the rest of this section, each one of these AE-based models is thoroughly described.

\subsection{Scorer}

%Scorer (\textbf{s}upervised \textbf{c}lass \textbf{o}ve\textbf{r}lap [rat\textbf{e}?] \textbf{r}eduction)

The first of our approaches to complexity reduction is to directly employ one of the complexity metrics as penalty, assuming that, if an AE is able to optimize this metric for a given dataset, the resulting representation will be less complex than the original. For this purpose, Fisher's discriminant ratio has been selected, as it is simple enough to be computed on the fly during training. The result is an AE which performs supervised class overlap reduction, or Scorer for short.

In order to introduce a complexity penalty based on Fisher's discriminant ratio, we consider the average of the discriminant ratios of each feature. This is different to the complexity measure commonly known as maximum Fisher's discriminant ratio or F1, which instead calculates the maximum of those ratios. In this case, we chose the average because it should provide better gradients in order to optimize the objective. This was corroborated by a preliminary experimentation.

The following equations formally define the complexity penalty computed within Scorer, $N^+$ denoting the amount of positive examples and $N^-$ the number of negative ones. First, we define the necessary terms for the mean of each variable for positive instances and the same for negative instances.
\begin{equation} \mu_j^+ = \frac 1 {N^+} \sum_{(x,+1)\in S} f(x)_j,\quad \mu_j^- = \frac 1 {N^-} \sum_{(x,-1)\in S} f(x)_j,\quad\end{equation}

Next comes the standard deviation for each variable and for each class, calculated as the mean squared value minus the square of the mean:
\begin{equation}\sigma_j^+ = \left[\frac 1 {N^+} \sum_{(x,+1)\in S} f(x)_j^2\right] - (\mu_j^+)^2 \end{equation}
\begin{equation}\sigma_j^+ = \left[\frac 1 {N^-} \sum_{(x,-1)\in S} f(x)_j^2\right] - (\mu_j^-)^2 \end{equation}

This allows to put together an expression for the average Fisher's discriminant ratio, which is introduced in the loss function in a way that ensures its value to be between 0 and 1:
\begin{equation}F=\frac{1}{n_f}\sum_{j=1}^{n_f} \frac{(\mu_j^+-\mu_j^-)^2}{\sigma_j^++\sigma_j^-},\quad\Omega(\theta;S)=\frac{1}{1 + F}\end{equation}

\begin{figure}[ht]
  \centering
  \includegraphics[width=.35\textwidth]{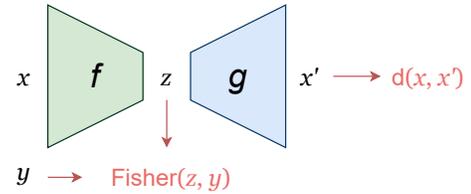}
  \caption{\label{fig.scorer}\change{Schematic illustration of the Scorer model. The average Fisher discriminant ratio of the encoded class distributions contributes to the training loss.}}
\end{figure}

The penalty term, drawing inputs from the encoding layer and the class label, is simply computed at the end of each training and added to the loss function, multiplied by a weight hyperparameter $\lambda$ so as to balance it with the reconstruction objective. Figure~\ref{fig.scorer} extends the basic AE diagram with these new components in order to illustrate how Scorer differs from the basic model.

\subsection{Skaler}

The next step in inducing a class-separating behavior in an AE is to use information theory-based measures. In this case, the AE is not forced to directly optimize a complexity metric. Instead, it receives information about the current relation among class distributions, and is assessed according to the similarity of those.

Although cross entropy is the conventional measure for classification loss, we refrain from using it as a penalty because the objective is not to directly classify, thus concentrating all instances on one of two points, but to provide a representation that better clusters examples.

\begin{figure}[ht]
  \centering
  \includegraphics[width=.35\textwidth]{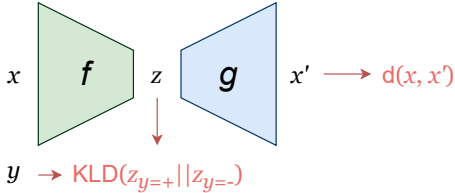}
  \caption{\label{fig.skaler}\change{Schematic illustration of the Skaler model. The KLD between the positive and the negative-class encodings contributes negatively to the training loss.}}
\end{figure}

Skaler is a supervised feature extraction model with a KLD-based penalty for class separation. As explained above, the KLD gives an asymmetric view on how two distributions are different. In this case, the objective is to maximize the difference among the distribution of encodings belonging to the positive class and those belonging to the negative class. A schematic view is provided in Figure~\ref{fig.skaler}.

Both positive and negative encodings can be modeled as following categorical distributions, if we assume that each feature in the encoding can have a high (1) and a low (0) state and that the highest feature is the one that matters. This is a simplification but it helps build a KLD-based formulation that is easy to implement and able to train successfully. Then, the sample space of each distribution consists of events associated to each feature, indicating whether that feature is the highest. If $P_+(j)$ denotes the probability that the $j$-th feature is highest for positive instances and $P_-(j)$ does the same for the negative class, the KLD-based penalty function would be as follows:

% DONE quitar paréntesis de logs
\begin{equation} \Omega(\theta;S)=- \sum_{j=1}^{d}P_+(j) \log\frac{P_+(j)}{P_-(j)}\end{equation}

% DONE explain how this is the probability of the categorical distribution

Now, in order to compute valid probabilities for each case of the categorical distribution out of the encoding generated by the AE, we take the mean of each variable in a vector and perform the softmax activation function, obtaining a vector of values summing 1, thus representing a distribution. The $j$-th component of that vector corresponds to the probability that the $j$-th feature is high for any given data sample:

\begin{equation} P_+(j)=\operatorname{softmax}\left(\frac{1}{N^{+}}\sum_{(x,+1)\in S}f(x)\right)_j \end{equation}

\begin{equation} P_-(j)=\operatorname{softmax}\left(\frac{1}{N^{-}}\sum_{(x,-1)\in S}f(x)\right)_j \end{equation}

Some preliminary tests revealed that it is easy for this penalty to force encodings onto a single class-dependent value for any inputs. It was observed, however, that maximizing the entropy of the encoding variables helped prevent this issue, so it is added as a negative term to the penalty in the implementation.

\subsection{Slicer}

The third proposal of this work goes a bit further than the two previous ones, since it incorporates not only a different penalty function, but also additional learnable parameters.

This alternative regularization is inspired on least-squares support vector machines (LSSVM) \cite{suykens1999least}. These models attempt to learn the hyperplane which best separates both classes, but the difference between them and traditional SVMs lies on the objective function. For both models, the linear (non-kernelized) version of the classifier optimizes parameters $w$ and $b$ of the hyperplane $w^Tx+b$. However, the functions that both models minimize are different. In the case of the LSSVM, the parameters are fitted to optimize the following expression:

\begin{equation}\frac 1 2 \left\Vert w\right\Vert^2+\frac \beta 2\sum_{(x,y)\in S}\left(1 - y\left(w^Tx+b\right)\right)^2\end{equation}

The idea behind our model is to find a representation which facilitates the task of fitting a linear classifier. The resulting model is an AE for supervised linear classifier error reduction, hereinafter called Slicer.

In order for the model to compute the linear classifier objective function, we add trainable weights $w$ and $b$ to the computation graph of the neural network. These are used to get the output of a linear SVM, allowing thus to train the SVM and use it as a penalty to modify the behavior of the encoder at the same time:
\begin{equation}\Omega(\theta;S)=\frac 1 2 \left\Vert w\right\Vert^2+\frac \beta 2\sum_{(x,y)\in S}\left(1 - y\left(w^Tf(x)+b\right)\right)^2~,\end{equation}
where $\beta$ is just a hyperparameter weighting the importance of the LSSVM loss, and $w$ and $b$ are updated by the model after each epoch, just like the rest of neural network weights. In this case, the value of $y$ is $1$ for the positive class and $-1$ for the negative one.

% DONE: explicar beta y dejar to bonico

\begin{figure}[ht]
  \centering
  \includegraphics[width=.4\textwidth]{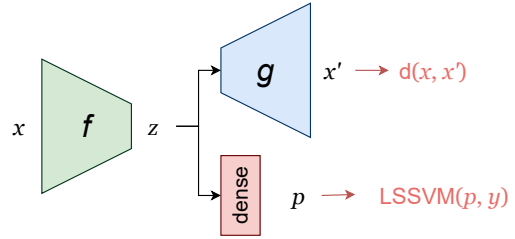}
  \caption{\label{fig.slicer}\change{Schematic illustration of the Slicer model. }}
\end{figure}

Figure~\ref{fig.slicer} illustrates how the AE is modified using the LSSVM objective function, taking the encoding $z=f(x)$ as input for the LSSVM and using a prediction $p=w^Tz+b$ to calculate its loss.

The result is a model that simultaneously trains a very simple classifier on the encoded data and uses its objective in order to find better representations. Our assumption is that there will exist some level of collaboration between both models and this will help the new features become more practical for classifiers to use.

\section{Experimental setup}\label{sec.setup}

Our proposals have been tested to verify their performance in reducing the complexity of data according to some measures, as well as in generating feature spaces where \rechange{binary} classification is easier. This section first goes through the materials for the experiments: data, methods and metrics, and then provides details on the implementations of the newly proposed models.

\iffalse \subsection{Experimental framework} \fi

The experiments that were performed in order to analyze whether using class information in an AE provides an advantage include a broad range of datasets as well as several well-established methods for comparison purposes. The following sections explain the data, compared methods and evaluation metrics used along the experimentation.

\subsection{Data}

The methods have worked with a collection of 27 datasets from several sources with varying dimensionalities.
\change{Thirteen of them originally have binary classes, six derive from the individual labels in a multilabel dataset, five are ``grouped'' binarizations where several classes are taken as the positive class and the rest as the negative one, and two originate from one-vs-all scenarios where only one arbitrary class is chosen as the positive one.}

\change{The grouped binarizations have been chosen so as to present a binary scenario that would make sense with each of the problems posed by the datasets: distinguishing vowels from consonants (when the original label was the letter), odd from even handwritten digits, walking from staying movement signals and two high-level categories of soil from images. One-vs-all schemes have not been performed in these cases due to the high amount of classes and resulting experiments.}

The datasets are briefly described and referenced \change{in the supplementary material.}

\subsection{Complexity reduction methods}

In addition to our proposals Scorer, Skaler and Slicer, we have selected four dimension reduction methods which can contribute to reducing the complexity of these datasets: PCA, LLE, Isomap and basic AE. PCA is used as the baseline for dimension reduction, LLE and Isomap are selected due to their manifold learning purpose, and the basic AE serves to analyze whether our proposals improve its behavior. None of these has the capability of learning from the classes, but they do address the dimensionality problem. The objective of our experiments is, thus, to test whether class information can be useful for an automatic feature learner to retrieve better quality attributes. Please refer to Table~\ref{tbl.methods} and Section~\ref{sec.methods} for a brief description of the idea behind each technique and a longer explanation, respectively.

\begin{table}[ht]
  \centering
  \caption{\label{tbl.methods}Brief description of each method available in the experiments}
  \begin{tabular}{ll}
    \toprule
    Method              & Description                                       \\
    \midrule
    PCA\cite{PCABook}   & Linear variance maximization                      \\
    LLE\cite{LLE}       & Neighborhood-based manifold learning              \\
    Isomap\cite{Isomap} & MDS-based manifold learning                       \\
    Basic AE\cite{AEs}  & Neural network for data reconstruction            \\
    Skaler              & Proposed AE with Kullback-Leibler-based penalty   \\
    Scorer              & Proposed AE with discriminant ratio-based penalty \\
    Slicer              & Proposed AE with LSSVM-based penalty              \\
    \bottomrule
  \end{tabular}
\end{table}

\subsection{Evaluation metrics}

In order to provide different perspectives on the performance of all methods, a diverse set of evaluation metrics has been selected. The objective is to be able to analyze the possible advantages and shortcomings of each available method.

We have trained our proposals and the compared methods to reduce the dimension of the datasets to the square root of the original dimension. The feature transformation learned by each one has been used to compute a list of complexity metrics for the resulting projections. Afterwards, we have trained several simple classifiers in order to assess the ease of classification with the generated features.

In summary, the metrics used for evaluation of each complexity reduction method can be categorized into classifier-agnostic and classifier-dependent.

% \paragraph{} 
Classifier-agnostic metrics are some of the complexity measures discussed in Section~\ref{sec.compmeas}. They have been chosen essentially for their popularity and easy interpretation. Morphology-based ONB metrics have also been computed so as to verify their affinity with the actual classification performance as well. %The list of metrics in this category is: maximum Fisher's discriminant ratio (F1), feature efficiency (F3), total and average number of balls ($\textit{ONB}_{\text{tot}}$ and $\textit{ONB}_{\text{avg}}$).

% \paragraph{} 
On the other hand, a partial objective of this experimentation is to check whether the generation of new features can actually ease classification tasks if aided by class information. The logical step is thus to analyze the performance of several datasets with the resulting variables.
\begin{itemize}
  \item F-score. Derived from precision (the ratio of instances correctly predicted as positive) and recall (the ratio of positive instances correctly detected), it is essentially the harmonic mean of both: \begin{equation}\text{F-score}=\frac{2\times\text{Precision}\times\text{Recall}}{\text{Precision}+\text{Recall}}\end{equation}
  \item Area under the ROC curve (AUC). This metric is computed as the area, out of 1, that lays under the receiver operating characteristic curve, which denotes, as the prediction threshold goes from 0 to 1, the ratio of true positives related to the ratio of false positives.
  \item Cohen's Kappa. It measures the level of agreement between the predictor and the ground truth, that is, the extent to which the coincidences differ from random chance ($p_c$). The mathematical definition \change{is}: \begin{equation}\kappa=1-\frac{1-\text{Acc}}{1-p_c}~.\end{equation}
\end{itemize}
These evaluation metrics have been chosen over other popular ones such as accuracy or precision since they attempt to provide a better overall picture of the performance without being affected by imbalance.
Some other metrics that are also considered common complexity measures are not actually classifier-independent: linear classifier error (L2), nonlinearity of linear classifier (L3) and 1NN classifier error (N3). These were previously defined in Section~\ref{sec.compmeas}.

Table~\ref{tbl.metrics} gathers all of these selected metrics with a short interpretation of each one. In order to obtain robust values, they have been computed over a 5-fold cross validation scheme.

\begin{table}[ht]
  \centering
  \caption{\label{tbl.metrics}Brief description of each evaluation metric used for the experiments, \change{classified according to their dependency on a classifier}}
  \begin{tabular}{lll}
    \toprule
     & Metric          & Meaning                                      \\
    \midrule
    \multirow{4}{*}{\rotatebox[]{90}{Agnostic}}
    %\multicolumn{2}{c}{Classifier-agnostic}\\\midrule
     & F1 (Fisher)     & Class overlap according to mean and variance \\
    % F2 (volume) & Volume of overlapping regions\\
     & F3 (efficiency) & Feature ability to separate classes          \\
     & ONB (total)     & Total number of balls in cover               \\
     & ONB (average)   & Average number of balls in cover             \\
    \midrule
    \multirow{6}{*}{\rotatebox[]{90}{Dependent}}
    % \multicolumn{2}{c}{Classifier-dependent}\\\midrule
     & L2              & Linear classifier error                      \\
     & L3              & Linear classifier nonlinearity               \\
     & N3              & 1NN classifier error                         \\
     & F-score         & Tradeoff between precision and recall        \\
     & AUC             & Area under the ROC curve                     \\
     & Kappa           & Agreement between prediction and truth       \\
    \bottomrule
  \end{tabular}
\end{table}

\subsection{Parameters}

The last details about the execution of the experiments are provided in Table~\ref{tbl.params}, which shows all the values for the parameters involved in the each of the different methods.

\begin{table}[ht]
  \centering
  \caption{\label{tbl.params}Enumeration of parameters used throughout the experiments.}
  \begin{tabular}{ll}
    \toprule
    Parameter                               & Value                                                            \\
    \midrule
    % \multicolumn{2}{c}{Classifier-independent}\\\midrule
    Encoding dimension                      & $\max\left\{\min\left\{\sqrt d, \frac{n}{10}\right\}, 2\right\}$ \\
    Epochs                                  & 200                                                              \\
    Number of hidden layers                 & 3 (1 for $<100$ variables)                                       \\
    Activation function (AEs except Skaler) & ReLU                                                             \\
    Activation function (Skaler)            & Sigmoid                                                          \\
    Penalty weight - Scorer                 & 0.01                                                             \\
    Penalty weight - Skaler                 & 0.1                                                              \\
    Penalty weight - Slicer                 & 1                                                                \\
    Reconstruction error                    & Cross entropy                                                    \\
    \bottomrule
  \end{tabular}
\end{table}

\iffalse The encoding dimension refers to the dimension of the middle layer in the case of AEs, and the selected dimension for the projection for the rest of methods. The square root of the original dimension of the dataset was chosen as it provides proportionally more reduction for higher-dimensional datasets.
  The number of epochs is the amount of training iterations that each autoencoder performs, passing the whole dataset through the network each time. The number of hidden layers is set to 3 in the general case, except for relatively small (less than 100 variables) datasets which are associated to a shallow AE. The activation function refers to the nonlinearity applied in the middle layer, the rest of hidden layers always use the ReLU function.

  Setting only one penalty weight for all datasets can be difficult since it is not optimal for every case, the weight for each model was selected so that the penalty was generally in the same order of magnitude or one order lower than the reconstruction error. This error was chosen to be cross entropy and each variable is min-max normalized to the $[0,1]$ interval in order for this distance to work properly.

\fi

\section{Experimental results}\label{sec.experiments}

This section presents the outcomes of the experiments performed, focusing on comparing the different methods, as well as drawing conclusions from the obtained metrics and graphics.

% DONE indicar que las abreviaturas no son mías, son de los papers originales
% DONE explicar la N3
% TODO homogeneizar títulos de tablas

\subsection{Results}

Experiments for the 27 datasets have been conducted in a 5-fold cross validation scheme. A total of 16 metrics were computed for each case, and the full results are available at the associated website\footnote{{https://ari-dasci.github.io/S-reducing-complexity/}}. Next, we show and analyze aggregated results and the corresponding statistical tests.

Table~\ref{tbl.ranking} holds the average ranking that each dimensionality reduction method achieved for each metric throughout the dataset collection. The winning method for each row is marked in underlined bold text. The number of overall first positions in rankings is summed up and shown in the last row of the table. The first observation that can be extracted is that model Slicer turns out to be consistently superior in most metrics, resulting in a vastly higher amount of won cases than the rest.

\begin{table}[ht]
  \caption{\label{tbl.ranking}Average ranking for each method in each evaluated metric. A horizontal line separates complexity metrics from classifier evaluation metrics. The best method is marked with a star \(\bigstar\). Those which are worse with $p<0.05$ are marked with $\times$, and those which are worse with $p<0.01$ are marked with $\otimes$. The total number of first positions achieved by each method is shown in the last row.}\centering
  \resizebox*{\linewidth}{!}{\change{\begin{tabular}{llrrrrrrr}
      \toprule
                                             &                             & PCA                        & LLE                & \scriptsize Isomap & AE                & \scriptsize Skaler                      & \scriptsize Scorer & \scriptsize Slicer \tabularnewline
      \midrule
      % \endhead
                                             & F1                          & \(\otimes\) 5.885          & \(\otimes\) 6.115  & \(\otimes\) 5.731  & \(\otimes\) 4.038 & \(\bigstar\) {\bfseries1.577} & 2.308              & 2.346 \tabularnewline
      % &F2 & 2.923 & \underline{\bfseries2.577} & 3.000 & 5.115 & 6.462 & 4.481 & 3.442 \tabularnewline
                                             & F3                          & \(\otimes\) 4.250          & \(\otimes\) 5.231  & \(\otimes\) 4.846  & \(\otimes\) 5.654 & 2.788                                   & 3.000              & \(\bigstar\) {\bfseries2.231} \tabularnewline
                                             & N3                          & \(\otimes\) 4.692          & \(\otimes\) 5.173  & \(\otimes\) 5.308  & \(\times\) 4.269  & 3.654                                   & 2.577              & \(\bigstar\) {\bfseries2.327} \tabularnewline
                                             & L2                          & \(\bigstar\) {\bfseries2.712} & \(\times\) 4.519   & 3.981              & \(\otimes\) 5.654 & \(\times\) 4.442                        & 3.846              & 2.846 \tabularnewline
                                             & L3                          & 2.769                      & \(\times\) 4.500   & 4.115              & \(\otimes\) 5.500 & \(\otimes\) 4.923                       & 3.654              & \(\bigstar\) {\bfseries2.538} \tabularnewline
                                             & $\textit{ONB}_{\text{tot}}$ & \(\otimes\) 4.481          & \(\otimes\) 6.115  & \(\otimes\) 4.519  & \(\otimes\) 4.481 & 3.442                                   & 3.019              & \(\bigstar\) {\bfseries1.942} \tabularnewline
                                             & $\textit{ONB}_{\text{avg}}$ & \(\otimes\) 4.442          & \(\otimes\) 6.077  & \(\otimes\) 4.519  & \(\otimes\) 4.577 & \(\times\) 3.538                        & 2.904              & \(\bigstar\) {\bfseries1.942} \tabularnewline\midrule
      \multirow{3}{*}{\rotatebox[]{90}{kNN}} & F-score                     & 3.096                      & \(\otimes\) 6.923  & \(\otimes\) 4.904  & 4.115             & 3.519                                   & 3.231              & \(\bigstar\) {\bfseries2.212} \tabularnewline
                                             & AUC                         & 3.288                      & \(\otimes\) 7.000  & \(\otimes\) 4.750  & \(\times\) 4.038  & 3.500                                   & 3.250              & \(\bigstar\) {\bfseries2.173} \tabularnewline\vspace{1ex}
                                             & Kappa                       & 3.096                      & \(\otimes\) 7.000  & \(\otimes\) 4.827  & \(\times\) 4.038  & 3.558                                   & 3.346              & \(\bigstar\) {\bfseries2.135} \tabularnewline
      \multirow{3}{*}{\rotatebox[]{90}{SVM}} & F-score                     & 3.788                      & \(\otimes\) 6.846  & \(\otimes\) 4.673  & 3.538             & 3.538                                   & 2.923              & \(\bigstar\) {\bfseries2.692}\tabularnewline
                                             & AUC                         & \(\times\) 4.000           & \(\otimes\) 6.846  & \(\otimes\) 4.865  & 3.462             & 3.346                                   & 2.962              & \(\bigstar\) {\bfseries2.519} \tabularnewline\vspace{1ex}
                                             & Kappa                       & 3.904                      & \(\otimes\)  6.846 & \(\otimes\) 4.827  & 3.346             & 3.577                                   & 2.962              & \(\bigstar\) {\bfseries2.538}\tabularnewline
      \multirow{3}{*}{\rotatebox[]{90}{MLP}} & F-score                     & \(\otimes\) 4.077          & \(\otimes\) 6.769  & \(\otimes\) 3.962  & \(\otimes\) 4.731 & 2.596                                   & \(\otimes\) 3.865  & \(\bigstar\) {\bfseries2.000} \tabularnewline
                                             & AUC                         & \(\otimes\) 4.000          & \(\otimes\) 7.000  & \(\otimes\) 4.058  & \(\otimes\) 4.635 & 2.404                                   & \(\otimes\) 3.942  & \(\bigstar\) {\bfseries1.962} \tabularnewline
                                             & Kappa                       & \(\otimes\) 4.000          & \(\otimes\) 7.000  & \(\otimes\) 4.077  & \(\otimes\) 4.596 & 2.558                                   & \(\otimes\) 3.827  & \(\bigstar\) {\bfseries1.942} \tabularnewline
      \midrule
                                             & wins                        & 62                         & 10                 & 17                 & 19                & 78                                      & 64                 & {\bfseries 188} \tabularnewline              % count overall wins better
      \bottomrule % DONE quitar F2, meter CDplot global
    \end{tabular}}}
\end{table}

Looking at the table into more detail, we can observe that the three supervised AE-based methods overall reach better metrics than the traditional feature extraction techniques, especially when analyzing the resulting predictive performance of the different classifiers.

In order to calculate which differences are significant, the Friedman's Aligned Ranks test %\cite{garcia2010advanced} 
was performed with its corresponding post-hoc test where the winning algorithm was chosen as the control for each metric. \change{Table~\ref{tbl.ranking}} organizes the results of these tests, indicating which methods  were significantly worse than the winner with two levels of confidence ($p<0.05$ and $p<0.01$). The tests allow to assess whether the values found by the rankings can be considered enough to state that two methods are performing differently. It is important to notice, however, that each statistical test had information only about a specific metric, and not the whole picture. As a result, the fact that a test does not find significant differences among two methods does not mean that they perform the same in general.

\begin{figure}[htbp]
  \centering
  \includegraphics[width=.9\linewidth]{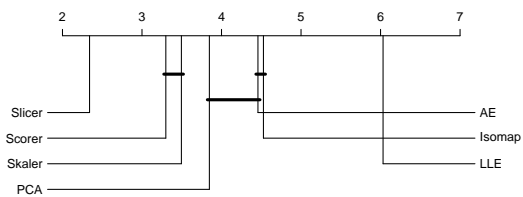}
  \caption{\label{fig.cdglobal}Critical distance plot for all results. Horizontal lines join methods where significant differences were not detected.}
\end{figure}

As a last summary, we have gathered all results and performed the Friedman's Aligned Ranks test in order to obtain potential global differences among methods. Although values from different metrics are mixed in these data, the test only considers rankings so it allows to extract some intuitions about the overall performance and whether different methods can be discerned from their evaluation metrics. This test is visualized in Figure~\ref{fig.cdglobal}, where critical distances are annotated with a confidence level of 99\% ($p<0.01$).

The supplementary material for this paper also includes density estimation plots which provide a visual account of the overall results of each method for each metric.

\subsection{Discussion}
The previous results allow to notice some interesting details. One of them is the fact that the model that performed best according to the F1 metric, Skaler, is not the one that attempts to optimize that metric. This suggests that, within a neural network model, the KLD-based loss penalty is more useful for this purpose than the F1 metric itself. It is also noteworthy that Skaler, having reached the best separability metrics according to the F1 metric, ended up losing performance to Slicer when evaluated by means of actual classifiers. This could mean that the F1 metric, despite being widely known and used, is not the best estimator of classifier performance.

Looking further at the relation between the complexity metrics and the classification metrics for each classifier, it is straightforward to identify the complexity metrics that align best with classifier performance. Those would be F3, N3, L3 and the morphology-based metrics. However, the complexity metrics where rankings are most similar to the classification rankings seem to be $\textit{ONB}_{\text{tot}}$ and $\textit{ONB}_{\text{avg}}$. %This can also be observed from the joyplots in Figures~\ref{fig.joyplotscomplexity} and~\ref{fig.joyplotsclassifiers}, which show similar, albeit horizontally flipped, shapes.

As to the comparison among complexity reduction methods, the computed rankings, critical distance plots and joyplots reveal that Slicer presents a clear advantage over the rest of methods in the majority of metrics. The individual statistical tests show many significant differences between Slicer and other methods, specifically LLE and Isomap. Some differences are also found from Slicer to PCA and AE, although not for every case. However, it is important to notice that Slicer is consistently superior to every other method across almost all metrics, something that these tests do not take into account. A more global perspective is given by the critical distance plot in Figure~\ref{fig.cdglobal}, which does find significant superiority of Slicer over the rest of methods.

In addition to Slicer taking the top place in classification tasks, we can also see that the other supervised AEs tend to be superior than the unsupervised methods in many cases, but the differences are smaller. In fact, PCA is also competing with them when the kNN classifier is employed for classification, which is interesting considering the simplicity of the method. The standard AE, however, does not achieve very good results in comparison to the improved versions with class-informed penalties. This leads to deduce that these types of regularizations are helping differentiate our proposals from what is otherwise the exact same neural network architecture. This idea is corroborated by the overall number of wins in Table~\ref{tbl.ranking}, where both Skaler and Scorer achieve a higher number of wins than the rest except Slicer.  Furthermore, the critical distance plot in Figure~\ref{fig.cdglobal} shows a significant difference from Scorer and Skaler to the classical feature extraction techniques, without being able to discard whether they perform equally.

In summary, the experimentation has shown that defining a class-based penalty in an otherwise unsupervised learning method such as an AE does help generate more useful features when the objective is training classifiers. Although the level of benefit will depend on the classifier used, any one of them will have some improvement in its performance. Of all the proposed approaches, the one that seems to retain the most information about classes within the encoded features is Slicer, which simultaneously learns the encoding and a linear classifier for the encoded variables.

\section{Final comments}\label{sec.comments}

\change{This work has introduced the concept of class complexity onto AEs that learn from class labels}. If dimension reduction helps classifiers by providing more compact versions of the data, complexity reduction goes further by also improving the shape and distribution of the different classes. \change{This concept} is realized in 3 specific models with distinct behavior, Scorer, Skaler and Slicer. 

\rechange{Similarly to other preprocessing methods, these intend to ease a later classification task.} In this case, only binary targets have been supported, since most of the complexity metrics were initially defined for binary problems. It would be possible to extend the proposed framework to multiclass or even multi-label datasets, although the penalty functions as well as the implementations would need notable modifications. A similar approach could also be followed for regression problems, where the target variable could help model the embedding space.

An additional option that may be explored is the application of a combination of penalties, since they are not necessarily exclusive, \change{including improvements on basic AEs such as variational AEs}. Some preliminary tests suggested that combining Slicer and Scorer may provide a slight improvement in classification performance, but it would need meticulous optimization of the selected penalties and their weights in order to provide promising results. \change{Further steps would include} introducing other complexity measures into the penalty function, so as to tackle other aspects of data complexity\change{, as well as learning from just a small number of labels in a semi-supervised scenario~\cite{kingma2014semi}}.

The experimentation developed to support the proposals has served not only to highlight the potential of Slicer as a better alternative for feature extraction, but also to notice which complexity measures are better predictors of classifier performance. In particular, morphology-based metrics like $\textit{ONB}_{\text{tot}}$ and $\textit{ONB}_{\text{avg}}$ seemed to be more aligned with classifiers.

\ifCLASSOPTIONcompsoc
  % The Computer Society usually uses the plural form
  \section*{Acknowledgments}
\else
  % regular IEEE prefers the singular form
  \section*{Acknowledgment}
\fi

We thank Jos{\'e} A. Fernández for suggesting the KLD as a possible separation measure. D. Charte is supported by the Spanish Ministry of Science under the FPU program (Ref.~FPU17/04069). F. Charte is supported by the Spanish Ministry of Science project PID2019-107793GB-I00 / AEI / 10.13039/501100011033. F. Herrera is supported by the Spanish Ministry of Science project PID2020-119478GB-I00  and the Andalusian Excellence project P18-FR-4961. This work is supported by the project DeepSCOP Ayudas Fundaci\'on BBVA a Equipos de Investigación Científica en Big Data 2018.

% The authors would like to thank Jos{\'e} Alberto Fern{\'a}ndez

% Can use something like this to put references on a page
% by themselves when using endfloat and the captionsoff option.
\ifCLASSOPTIONcaptionsoff
  \newpage
\fi

% trigger a \newpage just before the given reference
% number - used to balance the columns on the last page
% adjust value as needed - may need to be readjusted if
% the document is modified later
%\IEEEtriggeratref{8}
% The "triggered" command can be changed if desired:
%\IEEEtriggercmd{\enlargethispage{-5in}}

% references section

% can use a bibliography generated by BibTeX as a .bbl file
% BibTeX documentation can be easily obtained at:
% http://mirror.ctan.org/biblio/bibtex/contrib/doc/
% The IEEEtran BibTeX style support page is at:
% http://www.michaelshell.org/tex/ieeetran/bibtex/
\bibliographystyle{IEEEtran}

\begin{thebibliography}{10}
\providecommand{\url}[1]{#1}
\csname url@samestyle\endcsname
\providecommand{\newblock}{\relax}
\providecommand{\bibinfo}[2]{#2}
\providecommand{\BIBentrySTDinterwordspacing}{\spaceskip=0pt\relax}
\providecommand{\BIBentryALTinterwordstretchfactor}{4}
\providecommand{\BIBentryALTinterwordspacing}{\spaceskip=\fontdimen2\font plus
\BIBentryALTinterwordstretchfactor\fontdimen3\font minus
  \fontdimen4\font\relax}
\providecommand{\BIBforeignlanguage}[2]{{%
\expandafter\ifx\csname l@#1\endcsname\relax
\typeout{** WARNING: IEEEtran.bst: No hyphenation pattern has been}%
\typeout{** loaded for the language `#1'. Using the pattern for}%
\typeout{** the default language instead.}%
\else
\language=\csname l@#1\endcsname
\fi
#2}}
\providecommand{\BIBdecl}{\relax}
\BIBdecl

\bibitem{xiong2006enhancing}
H.~Xiong, G.~Pandey, M.~Steinbach, and V.~Kumar, ``Enhancing data analysis with
  noise removal,'' \emph{IEEE Transactions on Knowledge and Data Engineering},
  vol.~18, no.~3, pp. 304--319, 2006.

\bibitem{aggarwal2015outlier}
C.~C. Aggarwal, ``Outlier analysis,'' in \emph{Data mining}.\hskip 1em plus
  0.5em minus 0.4em\relax Springer, 2015, pp. 237--263.

\bibitem{AYESHA202044}
S.~Ayesha, M.~K. Hanif, and R.~Talib, ``Overview and comparative study of
  dimensionality reduction techniques for high dimensional data,''
  \emph{Information Fusion}, vol.~59, pp. 44--58, 2020.

\bibitem{ho2000measuring}
T.~K. Ho and M.~Basu, ``Complexity measures of supervised classification
  problems,'' \emph{IEEE transactions on pattern analysis and machine
  intelligence}, vol.~24, no.~3, pp. 289--300, 2002.

\bibitem{garcia2015data}
S.~Garc{\'\i}a, J.~Luengo, and F.~Herrera, \emph{Data preprocessing in data
  mining}.\hskip 1em plus 0.5em minus 0.4em\relax Springer, 2015, vol.~72.

\bibitem{Yang1999ARO}
Y.~Yang and X.~Liu, ``A re-examination of text categorization methods,'' in
  \emph{SIGIR '99}, 1999.

\bibitem{Dada2019MachineLF}
E.~Dada, J.~Bassi, H.~Chiroma, S.~Abdulhamid, A.~O. Adetunmbi, and O.~E.
  Ajibuwa, ``Machine learning for email spam filtering: review, approaches and
  open research problems,'' \emph{Heliyon}, vol.~5, 2019.

\bibitem{Russakovsky2015ImageNetLS}
O.~Russakovsky, J.~Deng, H.~Su, J.~Krause, S.~Satheesh, S.~Ma, Z.~Huang,
  A.~Karpathy, A.~Khosla, M.~S. Bernstein, A.~Berg, and L.~Fei-Fei, ``Imagenet
  large scale visual recognition challenge,'' \emph{International Journal of
  Computer Vision}, vol. 115, pp. 211--252, 2015.

\bibitem{deo2015machine}
R.~C. Deo, ``Machine learning in medicine,'' \emph{Circulation}, vol. 132,
  no.~20, pp. 1920--1930, 2015.

\bibitem{lorena2019complex}
A.~C. Lorena, L.~P. Garcia, J.~Lehmann, M.~C. Souto, and T.~K. Ho, ``How
  complex is your classification problem? a survey on measuring classification
  complexity,'' \emph{ACM Computing Surveys (CSUR)}, vol.~52, no.~5, pp. 1--34,
  2019.

\bibitem{weiss1995learning}
G.~M. Weiss, ``Learning with rare cases and small disjuncts,'' in \emph{Machine
  Learning Proceedings 1995}.\hskip 1em plus 0.5em minus 0.4em\relax Elsevier,
  1995, pp. 558--565.

\bibitem{DimRecComparative}
L.~Van Der~Maaten, E.~Postma, and J.~Van~den Herik, ``Dimensionality reduction:
  a comparative review,'' Tech. Rep., 2009.

\bibitem{bengio2013representation}
Y.~Bengio, A.~Courville, and P.~Vincent, ``Representation learning: A review
  and new perspectives,'' \emph{IEEE transactions on pattern analysis and
  machine intelligence}, vol.~35, no.~8, pp. 1798--1828, 2013.

\bibitem{charte-tutorial}
D.~Charte, F.~Charte, S.~Garc{\'\i}a, M.~J. del Jesus, and F.~Herrera, ``A
  practical tutorial on autoencoders for nonlinear feature fusion: Taxonomy,
  models, software and guidelines,'' \emph{Information Fusion}, vol.~44, pp.
  78--96, 2018.

\bibitem{ng2011sparse}
A.~Ng, ``Sparse autoencoder,'' \emph{CS294A Lecture notes}, vol.~72, no. 2011,
  pp. 1--19, 2011.

\bibitem{basubook}
M.~Basu and T.~K. Ho, \emph{Data complexity in pattern recognition}.\hskip 1em
  plus 0.5em minus 0.4em\relax Springer Science \& Business Media, 2006.

\bibitem{steinwart2008support}
I.~Steinwart and A.~Christmann, \emph{Support vector machines}.\hskip 1em plus
  0.5em minus 0.4em\relax Springer Science \& Business Media, 2008.

\bibitem{pascual2020revisiting}
J.~D. Pascual-Triana, D.~Charte, M.~A. Arroyo, A.~Fern{\'a}ndez, and
  F.~Herrera, ``Revisiting data complexity metrics based on morphology for
  overlap and imbalance: Snapshot, new overlap number of balls metrics and
  singular problems prospect,'' \emph{Knowledge and Information Systems}, 2021
  (forthcoming).

\bibitem{pcccd}
A.~Manukyan and E.~Ceyhan, ``Classification of {Imbalanced} {Data} with a
  {Geometric} {Digraph} {Family},'' \emph{Journal of Machine Learning
  Research}, 2016.

\bibitem{aggarwal2001surprising}
C.~C. Aggarwal, A.~Hinneburg, and D.~A. Keim, ``On the surprising behavior of
  distance metrics in high dimensional space,'' in \emph{International
  conference on database theory}.\hskip 1em plus 0.5em minus 0.4em\relax
  Springer, 2001, pp. 420--434.

\bibitem{zhang2013divergence}
Y.~Zhang, S.~Li, T.~Wang, and Z.~Zhang, ``Divergence-based feature selection
  for separate classes,'' \emph{Neurocomputing}, vol. 101, pp. 32--42, 2013.

\bibitem{wang2008feature}
L.~Wang, ``Feature selection with kernel class separability,'' \emph{IEEE
  Transactions on Pattern Analysis and Machine Intelligence}, vol.~30, no.~9,
  pp. 1534--1546, 2008.

\bibitem{quinlan}
J.~R. Quinlan, \emph{C4.5: Programs for Machine Learning}.\hskip 1em plus 0.5em
  minus 0.4em\relax San Francisco, CA, USA: Morgan Kaufmann Publishers, 1993.

\bibitem{PCA}
K.~Pearson, ``{LIII}. {On} lines and planes of closest fit to systems of points
  in space,'' \emph{Philosophical Magazine Series 6}, vol.~2, no.~11, pp.
  559--572, Nov 1901.

\bibitem{PCABook}
I.~T. Jolliffe, ``Introduction,'' in \emph{Principal component analysis}.\hskip
  1em plus 0.5em minus 0.4em\relax Springer, 1986, pp. 1--7.

\bibitem{LDA}
R.~A. Fisher, ``The statistical utilization of multiple measurements,''
  \emph{Annals of Human Genetics}, vol.~8, no.~4, pp. 376--386, 1938.

\bibitem{li2006efficient}
H.~Li, T.~Jiang, and K.~Zhang, ``Efficient and robust feature extraction by
  maximum margin criterion,'' \emph{IEEE transactions on neural networks},
  vol.~17, no.~1, pp. 157--165, 2006.

\bibitem{PCAandFA}
I.~T. Jolliffe, ``Principal component analysis and factor analysis,'' in
  \emph{Principal component analysis}.\hskip 1em plus 0.5em minus 0.4em\relax
  Springer, 1986, pp. 115--128.

\bibitem{borg2005modern}
I.~Borg and P.~J. Groenen, \emph{Modern multidimensional scaling: Theory and
  applications}.\hskip 1em plus 0.5em minus 0.4em\relax Springer Science \&
  Business Media, 2005.

\bibitem{Isomap}
J.~B. Tenenbaum, V.~De~Silva, and J.~C. Langford, ``A global geometric
  framework for nonlinear dimensionality reduction,'' \emph{science}, vol. 290,
  no. 5500, pp. 2319--2323, 2000.

\bibitem{LLE}
S.~T. Roweis and L.~K. Saul, ``Nonlinear dimensionality reduction by locally
  linear embedding,'' \emph{science}, vol. 290, no. 5500, pp. 2323--2326, 2000.

\bibitem{van2008visualizing}
L.~Van~der Maaten and G.~Hinton, ``Visualizing data using t-sne.''
  \emph{Journal of machine learning research}, vol.~9, no.~11, 2008.

\bibitem{suarez2018tutorial}
J.~L. Su{\'a}rez, S.~Garc{\'\i}a, and F.~Herrera, ``A tutorial on distance
  metric learning: Mathematical foundations, algorithms and software,''
  \emph{arXiv preprint arXiv:1812.05944}, 2018.

\bibitem{goldberger2004neighbourhood}
J.~Goldberger, G.~E. Hinton, S.~Roweis, and R.~R. Salakhutdinov,
  ``Neighbourhood components analysis,'' \emph{Advances in neural information
  processing systems}, vol.~17, pp. 513--520, 2004.

\bibitem{weinberger2009distance}
K.~Q. Weinberger and L.~K. Saul, ``Distance metric learning for large margin
  nearest neighbor classification.'' \emph{Journal of machine learning
  research}, vol.~10, no.~2, 2009.

\bibitem{NCMC}
T.~{Mensink}, J.~{Verbeek}, F.~{Perronnin}, and G.~{Csurka}, ``Distance-based
  image classification: Generalizing to new classes at near-zero cost,''
  \emph{IEEE Transactions on Pattern Analysis and Machine Intelligence},
  vol.~35, no.~11, pp. 2624--2637, 2013.

\bibitem{bengio2012deep}
Y.~Bengio, ``Deep learning of representations for unsupervised and transfer
  learning,'' in \emph{Proceedings of ICML workshop on unsupervised and
  transfer learning}, 2012, pp. 17--36.

\bibitem{johnson2016perceptual}
J.~Johnson, A.~Alahi, and L.~Fei-Fei, ``Perceptual losses for real-time style
  transfer and super-resolution,'' in \emph{European conference on computer
  vision}.\hskip 1em plus 0.5em minus 0.4em\relax Springer, 2016, pp. 694--711.

\bibitem{robbins1951stochastic}
H.~Robbins and S.~Monro, ``A stochastic approximation method,'' \emph{The
  annals of mathematical statistics}, pp. 400--407, 1951.

\bibitem{adam}
D.~Kingma and J.~Ba, ``Adam: A method for stochastic optimization,'' in
  \emph{International Conference on Learning Representations}, 2015.

\bibitem{backprop}
D.~E. Rumelhart, G.~E. Hinton, and R.~J. Williams, ``Learning representations
  by back-propagating errors,'' \emph{Nature}, vol. 323, no. 6088, pp.
  533--538, 1986.

\bibitem{CHARTE202093}
D.~Charte, F.~Charte, M.~J. {del Jesus}, and F.~Herrera, ``An analysis on the
  use of autoencoders for representation learning: Fundamentals, learning task
  case studies, explainability and challenges,'' \emph{Neurocomputing}, vol.
  404, pp. 93 -- 107, 2020.

\bibitem{suykens1999least}
J.~A. Suykens and J.~Vandewalle, ``Least squares support vector machine
  classifiers,'' \emph{Neural processing letters}, vol.~9, no.~3, pp. 293--300,
  1999.

\bibitem{AEs}
D.~H. Ballard, ``Modular learning in neural networks,'' in \emph{Proceedings of
  the Sixth National Conference on Artificial Intelligence - Volume 1}, ser.
  AAAI'87.\hskip 1em plus 0.5em minus 0.4em\relax AAAI Press, 1987, pp.
  279--284.

\bibitem{kingma2014semi}
D.~P. Kingma, S.~Mohamed, D.~J. Rezende, and M.~Welling, ``Semi-supervised
  learning with deep generative models,'' in \emph{Advances in neural
  information processing systems}, 2014, pp. 3581--3589.

\end{thebibliography}
\end{document}